\documentclass[letterpaper,10pt,conference]{ieeeconf}

\IEEEoverridecommandlockouts
\usepackage{cite}
\usepackage{amsmath,amssymb,amsfonts}
\usepackage{graphicx}
\usepackage{booktabs}
\usepackage{multirow}
\usepackage{xcolor}
\usepackage{subcaption}
\usepackage{xspace}
\usepackage[capitalise]{cleveref}
\usepackage{cuted}   
\usepackage{capt-of} 
\usepackage[font=small]{caption}

\newcommand{\method}{\textsc{RRDF}\xspace}
\newcommand{\bb}[1]{\mathbf{#1}}
\newcommand{\Vset}{\mathcal{V}}
\newcommand{\Sset}{\mathcal{S}}
\newcommand{\Zset}{\mathcal{Z}}
\newcommand{\Cset}{\mathcal{C}}
\newcommand{\Rset}{\mathcal{R}}

\newcommand{\actraw}{ACT\xspace}
\newcommand{\actours}{ACT-\method}
\newcommand{\best}[1]{\textbf{#1}}
\definecolor{draftred}{RGB}{155,35,35}

\usepackage{booktabs}
\usepackage{multirow}
\usepackage[table]{xcolor}

\newcommand{\taskrule}{%
  \arrayrulecolor{black!20}%
  \specialrule{0.35pt}{1.3pt}{1.3pt}%
  \arrayrulecolor{black}%
}

\begin{document}

\title{\LARGE\bf
Register-Routed Delayed Fusion: Rewiring Shortcut-Prone \\ Observation Fusion in Visuomotor Imitation}


\author{Jieting Long$^{1*}$ \and Weidong Cai$^{1}$ \and Weiming Zhi$^{1}$
\thanks{$^{*}$email: {\tt\small jieting.long@sydney.edu.au}.}%
\thanks{$^{1}$School of Computer Science, The University of Sydney, Australia.}
}

\maketitle


\begin{abstract}
Visuomotor imitation policies combine high-dimensional visual observations with compact signals such as proprioception, and their fusion topology determines when and through which tokens these streams interact. In dense token fusion, visual tokens may attend directly to compact tokens from the first encoder layer, allowing action-predictive compact cues to influence spatial visual representations early in their formation. We ask whether controlling this route improves visual responsiveness and policy behavior. We introduce Register-Routed Delayed Fusion (\method{}), which masks direct compact--visual attention and stages cross-modal interaction through a learned register workspace. Its isolate--collect--route schedule protects an early stream-separated prefix and later permits only register-mediated exchange, while compact conditioning remains available to the native action generator. Across five simulation tasks and three real-robot tasks, \method{} matches or improves dense ACT under nominal conditions. Appearance-shift evaluations on four simulation tasks and held-out-position evaluations on three real-robot tasks also favor RRDF. Phase-matched input probes show lower measured state-to-image sensitivity, while ablations indicate that adding registers alone does not reproduce the full performance gain. These results support controlling cross-modal propagation while retaining compact action conditioning.
\end{abstract}

\section{Introduction}
\label{sec:introduction}

Visuomotor imitation policies predict future robot actions from heterogeneous observation histories, typically combining high-dimensional visual inputs with compact signals such as robot proprioception and policy-internal context.  Recent work focused more on action generation, including action chunking ~\cite{zhao2023act,lee2024interact}, diffusion ~\cite{chi2023diffusionpolicy,Ze2024DP3,yang2024equibot,liu2024rdt}, and flow matching ~\cite{black2024pi0,zhang2025flowpolicy,jiang2025streaming}. We examine how layer-wise observation-fusion routes affect visual representations and closed-loop policy behavior.

We study this fusion stage through its layer-wise communication topology. Under dense token fusion, spatial visual and compact tokens participate in unrestricted self-attention from the first encoder layer, creating a direct compact-to-visual route, $\Cset\!\rightarrow\!\Vset$. Compact signals are often strongly correlated with task phase and future actions, so this one-hop route allows action-predictive compact cues to influence spatial visual representations from the beginning of encoder computation. More broadly, imitation policies are known to exploit nuisance signals correlated with expert actions, leading to causal confusion and brittle behavior under distribution shift~\cite{dehaan2019causal,park2021objectaware}. Recent work further shows that proprioceptive signals can dominate vision--proprioception policies and suppress visual utilization~\cite{lu2026when}. We hypothesize that unrestricted early compact-to-visual attention can encourage overreliance on action-correlated compact cues at the expense of task-relevant visual evidence. Consistent with this hypothesis, our intervention analysis later reveals greater sensitivity to state relative to image interventions under dense fusion, at both the representation and action-output levels.

\begin{figure}
    \centering
    \includegraphics[width=\linewidth]{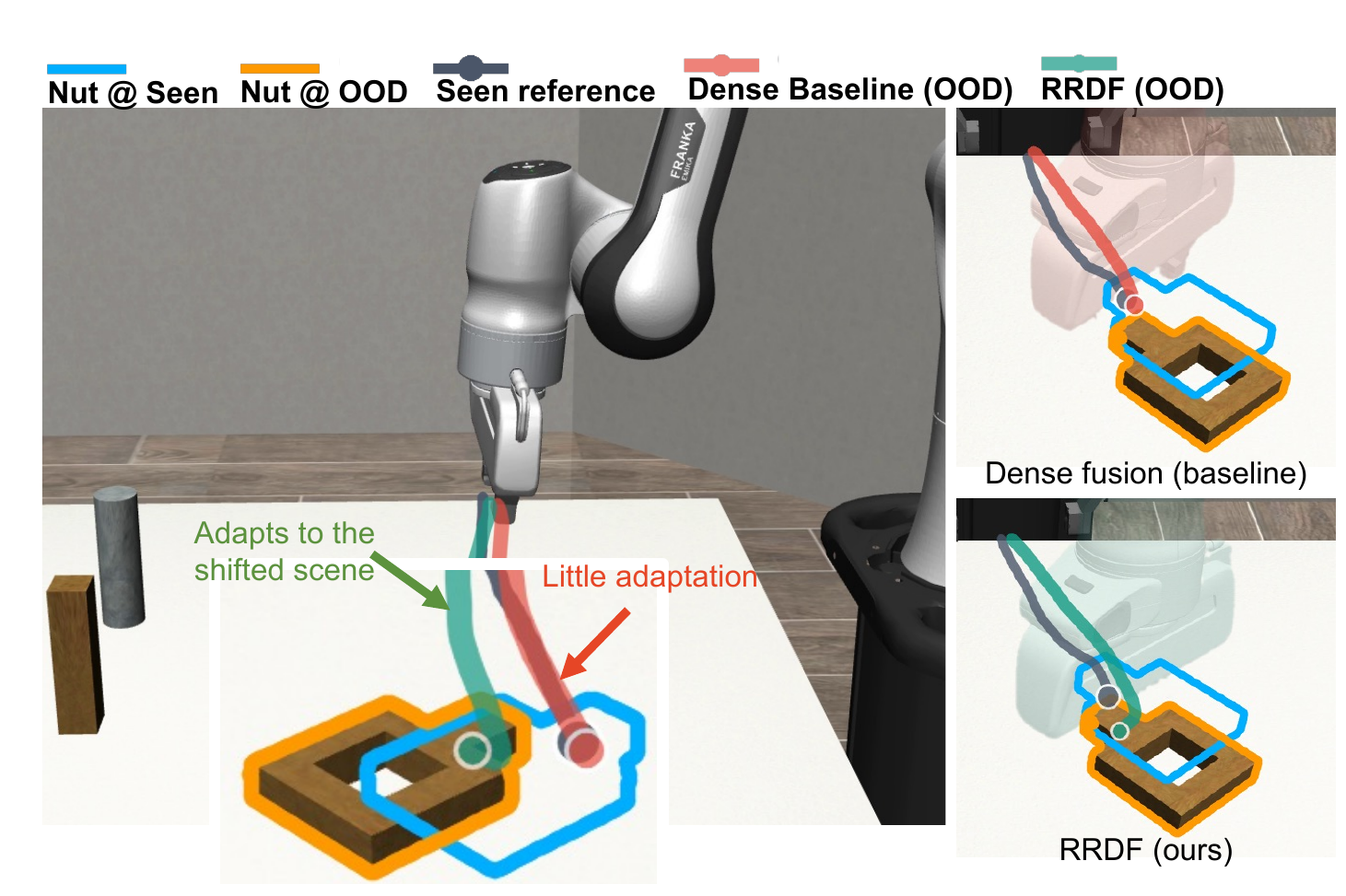}
\caption{\textbf{Representative behavioral difference under a held-out object-position shift.}
The dense-fusion policy produces similar trajectories for seen and shifted object configurations, failing to accommodate the shifted scene. \method{} instead adapts its trajectory and completes the task, illustrating improved scene responsiveness under the proposed delayed-fusion topology.}
    \vspace{-0.2em}
    \label{fig:teaser1}
\end{figure}

\begin{figure*}[t]
    \centering
    \includegraphics[width=\linewidth]{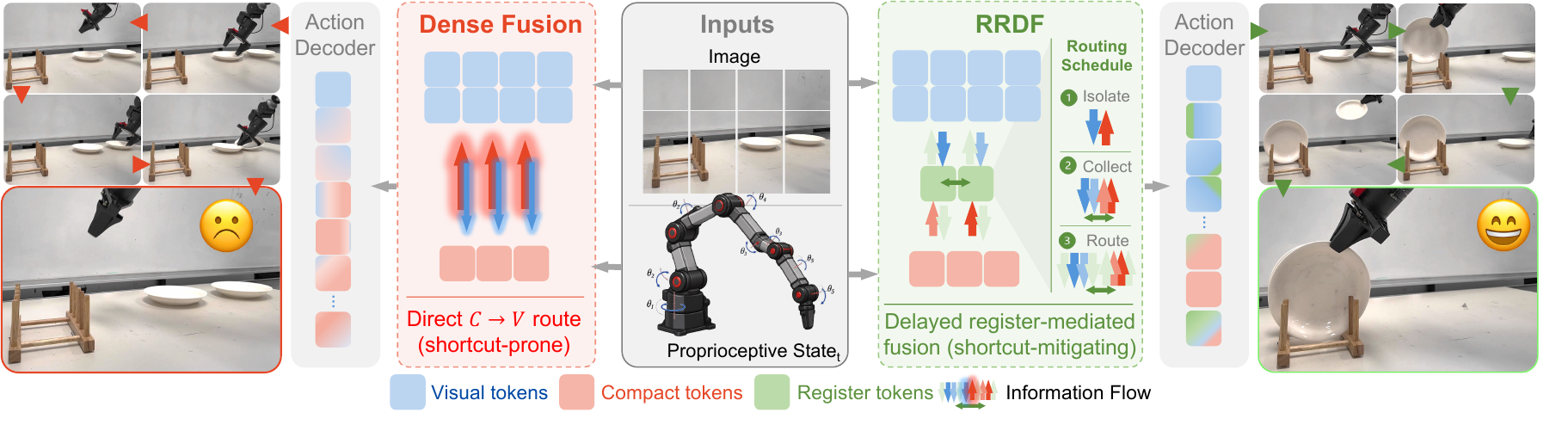}
    \captionof{figure}{\textbf{Overview of \method{}.} Dense fusion permits direct compact–visual attention from the first encoder layer. \method{} instead uses source-specific register accumulation, joint collection without primary-stream readback, and later register-mediated feedback. The outer panels show representative dense-fusion failure and \method{} success under a held-out-position shift.}
    \label{fig:teaser}
    \vspace{-1.7em}
\end{figure*}

Our focus is therefore not whether compact conditioning should be available, but when and through which routes it enters spatial visual representations. Proprioceptive feedback provides useful control information and remains directly accessible to the action generator. We ask whether controlling early compact-to-visual propagation can improve policy behavior without removing this conditioning. To address this issue, we introduce Register-Routed Delayed Fusion (\method{}, illustrated in \Cref{fig:teaser}), which masks direct compact--visual attention throughout the fusion encoder and delays cross-modal influence on the primary streams through a learned register workspace. Its isolate–collect–route schedule protects an early stream-separated prefix, collects cross-modal context in registers, and later exposes it only through register-mediated paths. Compact conditioning remains directly available to the native action generator. 

Across five simulation tasks, \method{} matches or improves the reported nominal performance of dense fusion. Appearance-shift tests and real-robot held-out-position evaluations also favor RRDF, with task-dependent gains. \Cref{fig:teaser1} illustrates a representative difference in trajectory adaptation. We examine these results through phase-matched input-sensitivity probes and fusion-topology ablations.

Concretely, this paper makes the following contributions:
\begin{itemize}
    \item We introduce a source-partitioned isolate–collect–route schedule that blocks direct compact–visual attention, jointly collects information before primary-stream readback, and preserves compact action conditioning.

    \item
    We evaluate nominal performance on five simulation and three real-robot tasks, appearance shifts on four simulation tasks, and held-out-position configurations on three real-robot tasks.    
    
    \item
    We characterize the permitted encoder dependencies, show that register capacity alone does not explain the gains, and reveal reduced state-to-image sensitivity through phase-matched probes.
\end{itemize}

\section{Related Work}
\label{sec:related_work}

\noindent\textbf{Multimodal observation fusion in visuomotor policies.}
Visuomotor policies also differ in how heterogeneous observations are combined before action generation. \textit{Feature-level approaches}, such as Diffusion Policy~\cite{chi2023diffusionpolicy}, encode observations into a shared conditioning representation, whereas \textit{token-based fusions} retain explicit modality tokens within a shared Transformer. ACT~\cite{zhao2023act}, HPT~\cite{wang2024scalinghpt}, and sensorimotor pre-training~\cite{radosavovic2023rpt} process heterogeneous robot inputs in shared Transformer token spaces, while broader systems incorporate visual, linguistic, and additional sensory inputs~\cite{rt12022arxivrt1,jiang2023vima,chen2022visuotactile,barron2025crossModal}. \textit{Structured token fusions} further constrain inter-group communication: Octo~\cite{ghosh2024octo} employs blockwise attention over task, observation, and readout tokens. \method{} targets explicit token fusion and treats the layer-wise routes between separately addressable modality groups as an architectural design variable.

\noindent\textbf{Shortcut learning and proprioceptive dominance.}
Behavioral cloning is susceptible to \emph{causal confusion}, where features correlated with expert actions become predictive shortcuts that fail under distribution shift~\cite{dehaan2019causal,Bica2021invariantcausal}, and to \emph{copycat effects}, where strongly action-correlated history becomes an easier predictor than task-relevant observations~\cite{wen2020fighting,chuang2022resolving}. More recently, shortcut learning has been identified as a limiting factor in large robot-policy generalization~\cite{xing2025shortcut}. Closest to our motivation, Lu \emph{et al.}~\cite{lu2026when} show that proprioception can dominate vision-proprioception learning and suppress visual utilization, degrading generalization. Existing remedies modify data or invariance objectives~\cite{dehaan2019causal,Bica2021invariantcausal}, state or observation design~\cite{zhao2025nostate,zhu2022viola}, regularization~\cite{park2021objectaware}, priming~\cite{wen2022fighting}, or optimization dynamics~\cite{lu2026when}. Complementary work studies visual representation learning and pretraining~\cite{nair2022r3m,radosavovic2023realworld,hansen2023visuomotor,zhi2025building}. Recent VLA studies~\cite{zhao2026should} examine state representation, history length, and injection location; \method{} retains compact action conditioning while controlling layer-wise communication within the observation encoder.

\noindent\textbf{Register and bottleneck tokens.}
Learned latent tokens provide compact computational interfaces in Transformers. Perceiver~\cite{jaegle2021perceiver} uses latent bottlenecks for general perception, while Perceiver-Actor~\cite{shridhar2023peract} adapts this paradigm to robotic manipulation. Outside robotics, Multimodal Bottleneck Transformers constrain audio-visual fusion through shared bottleneck tokens and introduce fusion at a configurable depth~\cite{nagrani2021mbt}. InterACT~\cite{lee2024interact} combines segment-wise encoding with cross-segment attention over CLS summaries while retaining original visual features at its output. ViT registers~\cite{darcet2024registers} provide learned computational slots for visual representation learning under unrestricted self-attention. \method{} uses a source-partitioned register workspace with an explicit read/write schedule. Source-specific accumulation without primary-stream readback is followed by joint collection and only then register-mediated feedback into the primary streams. Direct compact–visual attention remains masked throughout. This directed, depth-dependent schedule controls when information enters, mixes within, and leaves the workspace.

\section{Preliminaries: Fusion Topologies}
\label{sec:preliminaries}
\subsection{Multimodal Observation Tokenization}
\label{sec:tokenization}
For a unified analysis of the fusion architectures, we represent multimodal observations by spatial visual tokens and compact conditioning tokens. For an observation history containing $M$ camera views $\bb I_t^{1:M}$ and proprioceptive state $\bb s_t$, modality-specific encoders produce:
\begin{align}
\Vset_t^0
&=
\operatorname*{cat}_{m=1}^{M}
f_{\rm vis}(\bb I_t^m)
\in \mathbb R^{N_V\times d},
\nonumber\\
\Sset_t^0
&=
f_{\rm s}(\bb s_t)
\in \mathbb R^{N_S\times d},
\nonumber\\
\Cset_t^0
&=
[\Sset_t^0;\Zset_t^0]
\in \mathbb R^{N_C\times d},
\label{eq:observation_tokens}
\end{align}
where $\Vset_t^0$ denotes spatial visual tokens, $\Sset_t^0$ proprioceptive tokens, and $\Zset_t^0$ optional policy-internal conditioning tokens, with $N_C=N_S+N_Z$. In the configurations considered, $N_V\gg N_C$. For policies using observation histories, tokens from multiple time steps are concatenated analogously. The construction of $\Zset_t^0$ is policy-specific.

\begin{figure}[t]
    \centering
    \includegraphics[width=\linewidth]{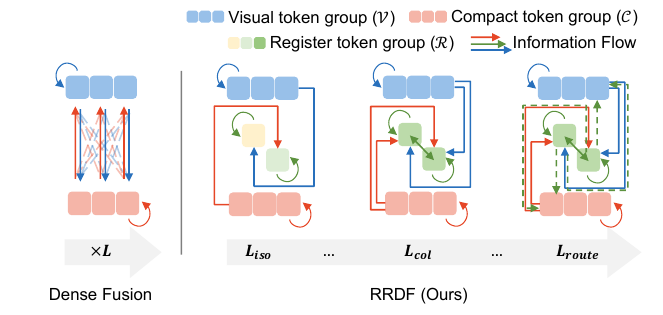}
    \vspace{-1em}
    \caption{\textbf{Depth-dependent directed topology of \method{}.} Left to right: dense fusion, source-specific register collection during Isolation, joint workspace formation during Collection, and register-mediated feedback during Routing. Direct $\mathcal C\leftrightarrow\mathcal V$  attention remains masked throughout; arrows denote permitted source-to-destination information flow.}
    \vspace{-0.3em}
    \label{fig:fusion_topologies}
\end{figure}

\subsection{Multimodal Fusion Topologies}
\label{sec:fusion_topologies}

A multimodal fusion topology specifies where the visual group $\Vset_t^0$ and compact group $\Cset_t^0$ interact and which information routes remain explicitly addressable.

\vspace{0.2em}
\noindent\textbf{Concatenation-based feature fusion:}
Modality-specific representations may first be independently aggregated and then concatenated into a shared observation condition: \begin{equation} \bb m_t = P\!\left[g_V(\Vset_t^0);g_C(\Cset_t^0)\right], \label{eq:feature_fusion} \end{equation}
where $g_V$ and $g_C$ produce modality-level features and $P$ is a learned projection. The resulting condition is consumed as a shared representation rather than as separately addressable modality-token groups.

\vspace{0.2em}
\noindent\textbf{Dense token fusion:}
Alternatively, modality-specific tokens may remain explicit and be processed jointly by a dense Transformer encoder: 
$E_{\rm dense}:$
\begin{align}
\bb X_t^0
&=
[\Cset_t^0;\Vset_t^0],
\nonumber\\
\bb X_t^\ell
&=
E_{\rm dense}^\ell(\bb X_t^{\ell-1}),
\qquad
\ell=1,\ldots,L,
\label{eq:dense_token_fusion}
\end{align}
where $E_{\rm dense}^\ell$ denotes the $\ell$-th Transformer encoder layer, whose unrestricted self-attention permits both within-group and cross-group communication. ACT is a representative instance of this topology ~\cite{zhao2023act}.

The communication topology at layer $\ell$ is represented by a routing matrix $\bb G^\ell$, whose rows index destination groups and columns index source groups: \begin{equation}
G_{d,s}^\ell=1
\quad\Longleftrightarrow\quad
s\!\rightarrow\!d,
\label{eq:topology_definition}
\end{equation}
where $s\!\rightarrow\!d$ means that queries in group $d$ may attend to keys and values in group $s$. Because communication may be directional and vary with depth, $\bb G^\ell$ need not be symmetric and is indexed by $\ell$.

Under the group order $(\Cset,\Vset)$, dense token fusion has: 
\begin{equation}
\bb G_{\rm dense}^\ell
=
\begin{bmatrix}
1 & 1\\
1 & 1
\end{bmatrix},
\qquad
\ell=1,\ldots,L.
\label{eq:dense_topology}
\end{equation}
Hence $G_{\Vset,\Cset}^\ell=1$ exposes the direct $\Cset\!\rightarrow\!\Vset$ route at every layer. \Cref{fig:fusion_topologies} contrasts this dense topology with the proposed depth-dependent \method{} topology.
\section{Register-Routed Delayed Fusion}
\label{sec:method}
Register-Routed Delayed Fusion (\method{}) restructures explicit multimodal token fusion through a source-partitioned register bank and depth-dependent attention masks. Direct $\Cset\leftrightarrow\Vset$ attention remains masked throughout the encoder: the primary streams cannot read the register workspace during the protected prefix, and later cross-modal exchange occurs only through register-mediated routes.


\subsection{Register-Augmented Fusion Encoder}
\label{sec:register_encoder}

\begin{figure*}[t]
\centering
\begin{subfigure}{0.6\textwidth}
    \includegraphics[width=\linewidth]{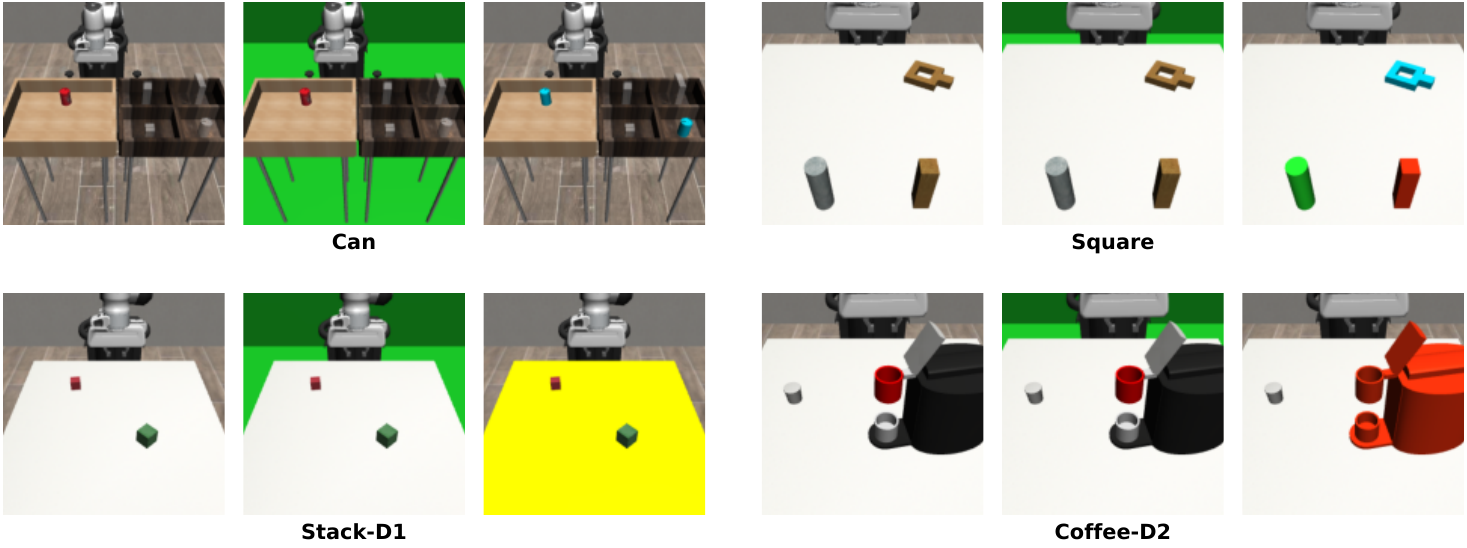}
    \caption{Representative observations and visual shifts}
    \label{fig:sim_stress_a}
\end{subfigure}
\hfill 
\begin{subfigure}{0.39\textwidth}
    \includegraphics[width=\linewidth]{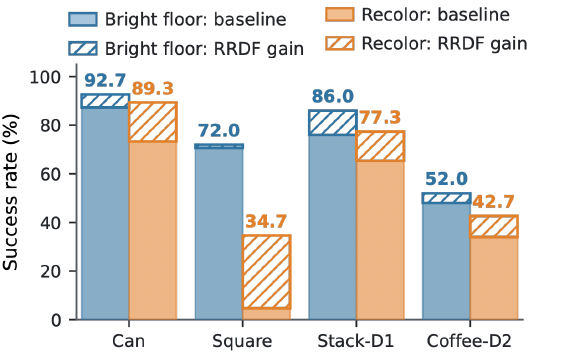}
    \caption{Quantitative results}
    \label{fig:sim_stress_b}
\end{subfigure}
\caption{\textbf{Controlled appearance-shift stress tests in simulation.}
(a) Nominal, bright-floor, and recolored observations for each task (left to right); geometry and dynamics are unchanged.
(b) Closed-loop success under each shift. Solid bars show dense-fusion success, hatched extensions the \method{} gain, and total height the \method{} result.}
\vspace{-1.5em}
\label{fig:sim_stress}
\end{figure*}

\vspace{0.2em}\noindent\textbf{Register workspace:}
A bank of $R$ learned register tokens~\cite{darcet2024registers} is appended to the fusion-encoder
sequence and partitioned into compact-facing and visual-facing
subsets:
\begin{align}
\Rset^0
&=
[\Rset_C^0;\Rset_V^0]
=
[\bb r_1,\ldots,\bb r_R]
\in\mathbb R^{R\times d},
\nonumber\\
\bb X_t^\ell
&=
[\Cset_t^\ell;
 \Rset_{C,t}^\ell;
 \Rset_{V,t}^\ell;
 \Vset_t^\ell],
\qquad
\ell=0,\ldots,L,
\label{eq:register_bank}
\end{align}
where
$\Rset_{C,t}^0=\Rset_C^0$ and
$\Rset_{V,t}^0=\Rset_V^0$
are independently learned initial embeddings. During isolation, the partition enforces source-specific access: $\Rset_C$ attends only to compact tokens and compact-facing registers, while $\Rset_V$ attends only to visual tokens and visual-facing registers. The partition defines routing permissions without assigning fixed semantic roles. From collection onward, both subsets share the same routing permissions and form the unified workspace $\Rset_t^\ell=[\Rset_{C,t}^\ell;\Rset_{V,t}^\ell]$.

The registers serve as learned internal encoder states rather than observations, action queries, or prediction targets, forming a communication workspace maintained across encoder depth. By contrast, adding the same registers under unrestricted self-attention leaves the direct compact-visual routes intact and serves as a capacity-matched control.

\vspace{0.2em}\noindent\textbf{Encoder interface:}
The register-routed encoder produces the final conditioning memory:
\begin{align}
\bb M_t^{\method}
&=
[\Cset_t^L;
 \Rset_{C,t}^L;
 \Rset_{V,t}^L;
 \Vset_t^L]
\nonumber\\
&=
E_{\phi}^{\method}
([\Cset_t^0;\Rset_C^0;\Rset_V^0;\Vset_t^0]).
\label{eq:rrdf_encoder}
\end{align}
The resulting memory is supplied through the action generator’s native conditioning interface, without modifying its architecture. Both compact tokens and register subsets remain available as conditioning tokens. Relative to dense fusion, \method{} therefore changes the encoder communication schedule and the resulting conditioning memory, rather than removing compact information from the policy.

\subsection{Delayed Fusion via Isolate-Collect-Route}
\label{sec:delayed_fusion}

The group-level topology is implemented via an additive attention mask~\cite{vaswani2017attention}. For
a query token $i$ and key token $j$ at layer $\ell$:
\begin{align}
M_{ij}^\ell
=
\begin{cases}
0, & G_{g(i),g(j)}^\ell=1,\\
-\infty, & G_{g(i),g(j)}^\ell=0,
\end{cases}
\label{eq:routing_mask}
\end{align}
where $g(i)\in\{\Cset,\Rset_C,\Rset_V,\Vset\}$ denotes the token group. The routing mask
is added to the attention logits and combined with any native padding,
temporal, or causal mask.

Let $L_{\rm iso}$ and $L_{\rm col}$ denote the numbers of isolation and
collection layers. The depth-dependent topology is:
\begin{align}
\bb G^\ell
=
\begin{cases}
\bb G_{\rm iso},
&1\leq\ell\leq L_{\rm iso},\\
\bb G_{\rm col},
&L_{\rm iso}<\ell\leq L_{\rm iso}+L_{\rm col},\\
\bb G_{\rm route},
&L_{\rm iso}+L_{\rm col}<\ell\leq L,
\end{cases}
\label{eq:routing_schedule}
\end{align}
with
$L=L_{\rm iso}+L_{\rm col}+L_{\rm route}$,
$L_{\rm col}\geq1$, and $L_{\rm route}\geq1$,
so that jointly collected register states can be consumed by a subsequent routing layer.

The three stage-specific topology matrices use the group order $(\Cset,\Rset_C,\Rset_V,\Vset)$, with rows indexing query groups and columns indexing source key-value groups. They define a progression from source-specific accumulation during \textit{Isolation}, through joint workspace formation during \textit{Collection}, to mediated redistribution during
\textit{Routing}:
\begingroup
\setlength{\arraycolsep}{1.7pt}
\renewcommand{\arraystretch}{0.9}
\begin{equation*}
\bb G_{\rm iso}=
\left[\begin{smallmatrix}
1&0&0&0\\
1&1&0&0\\
0&0&1&1\\
0&0&0&1
\end{smallmatrix}\right],\;
\bb G_{\rm col}=
\left[\begin{smallmatrix}
1&0&0&0\\
1&1&1&1\\
1&1&1&1\\
0&0&0&1
\end{smallmatrix}\right],\;
\bb G_{\rm route}=
\left[\begin{smallmatrix}
1&1&1&0\\
1&1&1&1\\
1&1&1&1\\
0&1&1&1
\end{smallmatrix}\right].
\end{equation*}
\endgroup

\begin{figure*}[t]
\centering
\includegraphics[width=\textwidth]{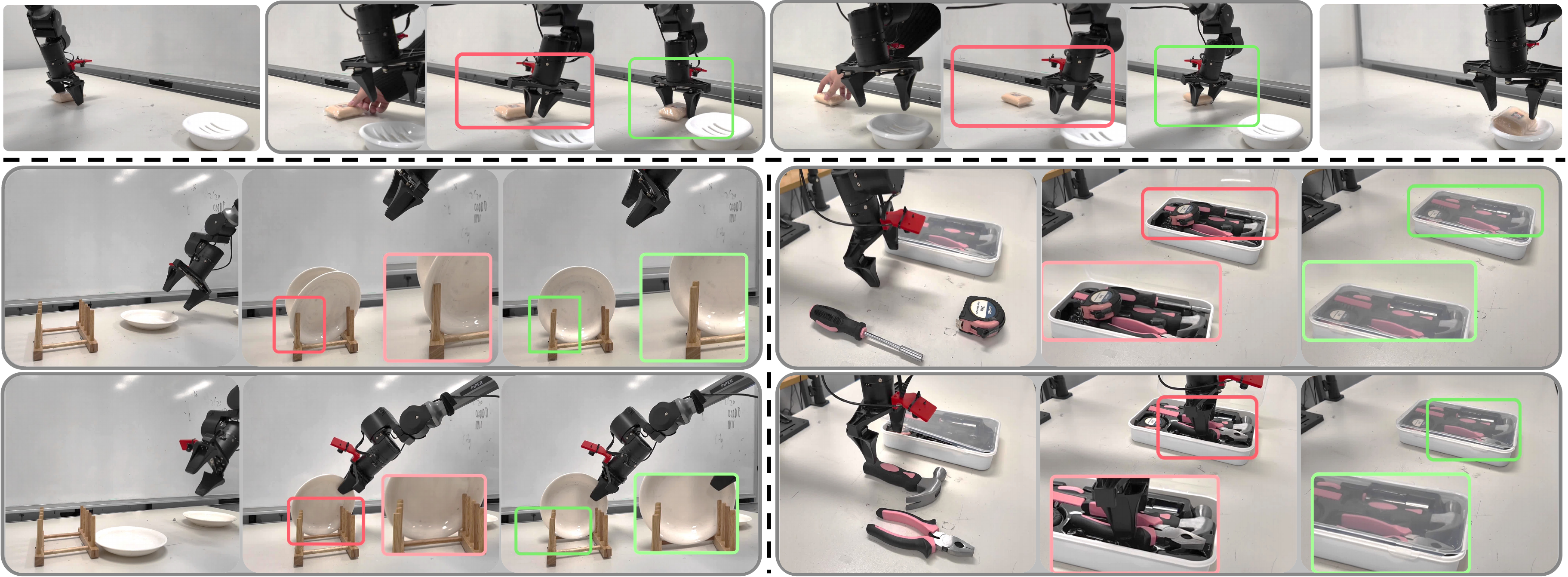}
\vspace{-1em}
\caption{\textbf{Qualitative real-robot behaviours.}
Matched dense-fusion and \method{} rollouts illustrate \textit{Spatial Adaptation} on Soap Pick-up (top), \textit{Precision Alignment} on Plate Stacking (bottom-left), and \textit{Precision Placement} on Toolbox (bottom-right), under nominal and held-out-position conditions. Red/green boxes mark representative dense-fusion failures and \method{} successes.}
\vspace{-1.5em}
\label{fig:qualitative}
\end{figure*}

\vspace{0.2em}\noindent\textbf{Isolate:}
The primary streams remain self-only, while $R_C$ and $R_V$ attend only to their corresponding source stream and register subset; neither primary stream reads the workspace.

\vspace{0.2em}\noindent\textbf{Collect:}
The source-specific restrictions are lifted within the register workspace. Both register subsets may attend to both primary streams and the full register bank, while the primary streams remain self-only. Because Transformer layers update synchronously, a dedicated collection layer lets the registers aggregate both streams before primary-stream readback.

\vspace{0.2em}\noindent\textbf{Route:}
The primary streams may now attend to themselves and the complete register bank, while the registers continue to collect from both streams. Direct $\Cset\!\leftrightarrow\!\Vset$ attention remains masked, so subsequent cross-modal interaction is mediated exclusively through the shared register workspace rather than restored through dense token fusion.

The routing delay \(L_{\rm delay}=L_{\rm iso}+L_{\rm col}\) controls when jointly informed register states become accessible to the primary streams, while \(R=R_C+R_V\) controls the mediated-interface width. The default topology mediates communication in both directions.


\subsection{Structural Properties of the Fusion Topology}
\label{sec:properties}

The routing schedule defines the admissible computational graph of the fusion encoder: a masked edge excludes the corresponding dependency, whereas an unmasked edge permits, but does not require, it.

\vspace{0.2em}\noindent\textbf{Exact direct-edge exclusion:}
The direct compact-visual entries remain masked at every fusion-encoder layer:
\begin{align}
G_{\Vset,\Cset}^{\ell}
=
G_{\Cset,\Vset}^{\ell}
=
0,
\qquad
1\leq\ell\leq L.
\label{eq:direct_route_exclusion}
\end{align}
The corresponding attention blocks are identically zero, so direct cross-group edges are excluded architecturally rather than merely down-weighted.

\begin{table}[t]
\centering
\caption{\textbf{Quantitative results in simulation.}
Performance denotes success rate or Push-T score. Entries show best-of-final-ten / final-ten summaries; statistical definitions follow \Cref{sec:simulation}.}
\label{tab:sim}
\footnotesize
\renewcommand{\arraystretch}{1.1}
\resizebox{\columnwidth}{!}{%
\begin{tabular}{@{}llcccc@{}}
\toprule
Task & Action & Perf. (\%) $\uparrow$ & SD. & Time (s) $\downarrow$ & SD. \\
\midrule
\multicolumn{6}{@{}l}{\textbf{ACT Baseline}} \\
\taskrule
Can
    & joint & \best{99.33}/96.67 & 1.15/1.44
    & 7.08/7.08 & 0.08/0.02 \\
Square
    & joint & 75.33/73.53 & 3.06/2.78
    & 10.56/10.49 & 0.68/0.15 \\
Stack-D1
    & EEF & 78.67/66.33 & 2.31/4.92
    & \best{6.11/6.17} & 0.43/0.27 \\
Coffee-D2
    & EEF & 51.33/37.07 & 9.45/5.98
    & \best{11.93/11.50} & 0.22/0.26 \\
Push-T & 2D & 77.61/72.76 & 34.28/13.73
                 & 22.72/23.06 & 7.79/7.37 \\
\midrule
\multicolumn{6}{@{}l}{\textbf{\method}} \\
\taskrule
Can
    & joint & \best{99.33/98.33} & 1.15/0.72
    & \best{6.97/7.03} & 0.19/0.03 \\
Square
    & joint & \best{93.33/86.13} & 3.06/2.70
    & \best{9.74/10.18} & 0.25/0.17 \\
Stack-D1
    & EEF & \best{88.67/73.73} & 6.11/5.32
    & 6.16/6.92 & 0.47/0.61 \\
Coffee-D2
    & EEF & \best{55.33}/\best{44.33} & 8.08/5.03
    & 12.05/11.60 & 0.47/0.30 \\
Push-T
    & 2D
    & \best{84.09/81.27}
    & 27.64/7.84
    & 23.06/23.61
    & 8.76/8.00 \\
\bottomrule
\end{tabular}%
}
\end{table}

\vspace{0.2em}\noindent\textbf{Visual-prefix computational independence:}
For fixed model parameters, group-separable input tokenizers, and standard Transformer blocks in which masked self-attention is the only cross-token operation, the visual-token states are computationally independent of the current compact input throughout the isolation and collection prefix:

\begin{align}
\frac{\partial\Vset_t^k}{\partial\Cset_t^0}
=
\bb 0,
\qquad
1\leq k\leq L_{\rm iso}+L_{\rm col}.
\label{eq:prefix_independence}
\end{align}
During isolation, compact and visual information are separated within the two register subsets. During collection, register states may depend on both streams, but the visual stream cannot access them until routing begins. This is a forward-input independence property at fixed parameters, not statistical independence or a guarantee against shortcut learning; visual tokens may still encode robot configuration visible in the images.

\vspace{0.2em}\noindent\textbf{Register-mediated paths:}
Because the direct compact-visual edges remain masked, cross-modal influence within the encoder must pass through at least one register state. In particular, compact information present in $\Cset_t^{\ell-1}$ can reach the visual stream through:
\begin{align}
\Cset_t^{\ell-1}
\longrightarrow
\Rset_t^\ell
\longrightarrow
\Vset_t^{\ell+1},
\label{eq:mediated_path}
\end{align}
whenever layer $\ell+1$ is a routing layer. The earliest such dependency enters the visual stream $\Vset$ at $L_{delay}$+1; the reverse path is analogous.



\section{Experiments and Results}
\label{sec:experiments}
We evaluate three aspects of the proposed fusion topology:
task performance under nominal and shifted observations,
representation- and action-level input sensitivity, and the effects of register capacity and communication schedules. These experiments assess the combined routing intervention; the sensitivity probes characterize responses to specified input changes rather than directly identifying shortcut use.

\vspace{0.2em}\noindent\textbf{Policy instantiation:}
We instantiate \method{} in ACT~\cite{zhao2023act} and compare against dense fusion, matching all non-fusion modules, data, and optimization settings. The compact group $C$ contains the proprioceptive state and policy-latent tokens. The four-layer encoder, by default, uses $(L_{\rm iso},L_{\rm col},L_{\rm route})=(1,1,2)$, with $R=2$ registers in simulation and $R=4$ on the real robot, evenly split between $R_C$ and $R_V$. The action generator remains unchanged.

\begin{table}[t]
\centering
\caption{\textbf{Quantitative results on real robots.}
Success rate (SR, \%): observed success rate $\pm$ estimated binomial standard error. Time: mean $\pm$ sample SD across successful rollouts; ``--'' denotes no success.}
\label{tab:real}
\footnotesize
\renewcommand{\arraystretch}{1.1}
\resizebox{\columnwidth}{!}{%
\begin{tabular}{@{}lcccc@{}}
\toprule
& \multicolumn{2}{c}{Nominal (ID)}
& \multicolumn{2}{c}{Held-out Position (OOD)} \\
\cmidrule(lr){2-3} \cmidrule(l){4-5}
Task
& SR (\%) $\uparrow$ & Time (s) $\downarrow$
& SR (\%) $\uparrow$ & Time (s) $\downarrow$ \\
\midrule
\multicolumn{5}{@{}l}{\textbf{\actraw}} \\
\taskrule
Soap Pick-up
& $55.0\pm11.1$ & $12.6\pm2.9$
& $30.0\pm10.2$ & $11.0\pm3.0$ \\

Plate Stacking
& $65.0\pm10.7$ & $28.5\pm3.5$
& $30.0\pm10.2$ & $29.5\pm4.0$ \\

Toolbox
& $50.0\pm11.2$ & $35.5\pm3.5$
& ${0.0 \pm 0.0}^{\dagger}$ & -- \\

\midrule
\multicolumn{5}{@{}l}{\textbf{\actours}} \\
\taskrule
Soap Pick-up
& $\mathbf{{100.0 \pm 0.0}^{\dagger}}$ & $12.1\pm2.6$
& $\mathbf{95.0\pm4.9}$ & $11.0\pm3.9$ \\

Plate Stacking
& $\mathbf{{100.0 \pm 0.0}^{\dagger}}$ & $27.8\pm2.5$
& $\mathbf{{100.0 \pm 0.0}^{\dagger}}$ & $29.0\pm3.3$ \\

Toolbox
& $\mathbf{85.0\pm8.0}$ & $34.0\pm4.0$
& $\mathbf{20.0\pm8.9}$ & $35.0\pm4.5$ \\
\bottomrule
\end{tabular}%
}
\end{table}
\begingroup
\renewcommand{\thefootnote}{\fnsymbol{footnote}}
\footnotetext[2]{%
At 0/20 or 20/20 successes, zero plug-in standard error does not imply
certainty about the true success probability.%
}
\endgroup

\subsection{Simulated Experiments}
\label{sec:simulation}

\vspace{0.2em}\noindent\textbf{Benchmarks and setup:}
We evaluate on RoboMimic Can and Square~\cite{robomimic2021}, MimicGen Stack-D1 and Coffee-D2~\cite{mandlekar2023mimicgen}, and Push-T~\cite{chi2023diffusionpolicy}, using joint-space, end-effector, and planar 2D actions, respectively. Performance is measured by success rate on the four manipulation tasks and the standard Push-T score, defined as the maximum normalized target-coverage reward attained during a rollout. Completion time uses successful rollouts only.

For the four manipulation tasks, each checkpoint uses three matched evaluation seeds with 50 rollouts each; success and successful-rollout completion time are reported as mean $\pm$ sample SD across seed-level summaries. Push-T uses 150 scene-initialization seeds with one rollout each; score and completion time are reported as mean $\pm$ sample SD across all rollouts and the successful subset, respectively. ``Best'' is the highest mean nominal performance among the final ten checkpoints; final-ten results are mean $\pm$ sample SD across those ten checkpoint means.

\vspace{0.2em}\noindent\textbf{Nominal performance:}
\Cref{tab:sim} shows that \method{} matches or improves dense fusion at the selected best checkpoint and also improves every final-ten-checkpoint average. Gains are largest on Square and Stack-D1, while the longer-horizon Coffee-D2 remains challenging for both.

\vspace{0.2em}\noindent\textbf{Controlled visual stress tests:}
We further evaluate RoboMimic~\cite{robomimic2021} Can and Square and MimicGen~\cite{mandlekar2023mimicgen} Stack-D1 and Coffee-D2 under rendering changes absent from the training demonstrations. The nominally selected best checkpoint is fixed across both
appearance shifts. Task dynamics and rollout initializations are matched across variants, and only rendering is modified; subsequent states evolve under each policy’s closed-loop actions. \Cref{fig:sim_stress} illustrates the perturbations and reports the corresponding closed-loop performance. \method{} achieves higher observed success in all eight evaluated task–appearance-shift conditions.

\subsection{Real-World Experiments}
\label{sec:real_world}

\vspace{0.2em}\noindent\textbf{Robot tasks and setup:}
Real-world experiments use a Piper-X manipulator with joint-space absolute control and third-person and egocentric views. We collect 24 demonstrations per task from predefined object placements within the robot’s reachable tabletop workspace. Held-out-position trials use placements excluded from demonstration collection, including intermediate locations between demonstrated positions and unseen joint object configurations. For Plate Stacking, demonstration collection fixes one plate while varying the other, whereas held-out trials reposition both plates. Each task-specific policy is evaluated at the final checkpoint
after 20,000 optimization steps, using 20 matched rollouts per
task and condition. Success rates are reported with estimated
binomial standard errors, and completion times as mean $\pm$
sample SD across successful rollouts.


\begin{figure}[t]
    \centering
    \includegraphics[width=\linewidth]{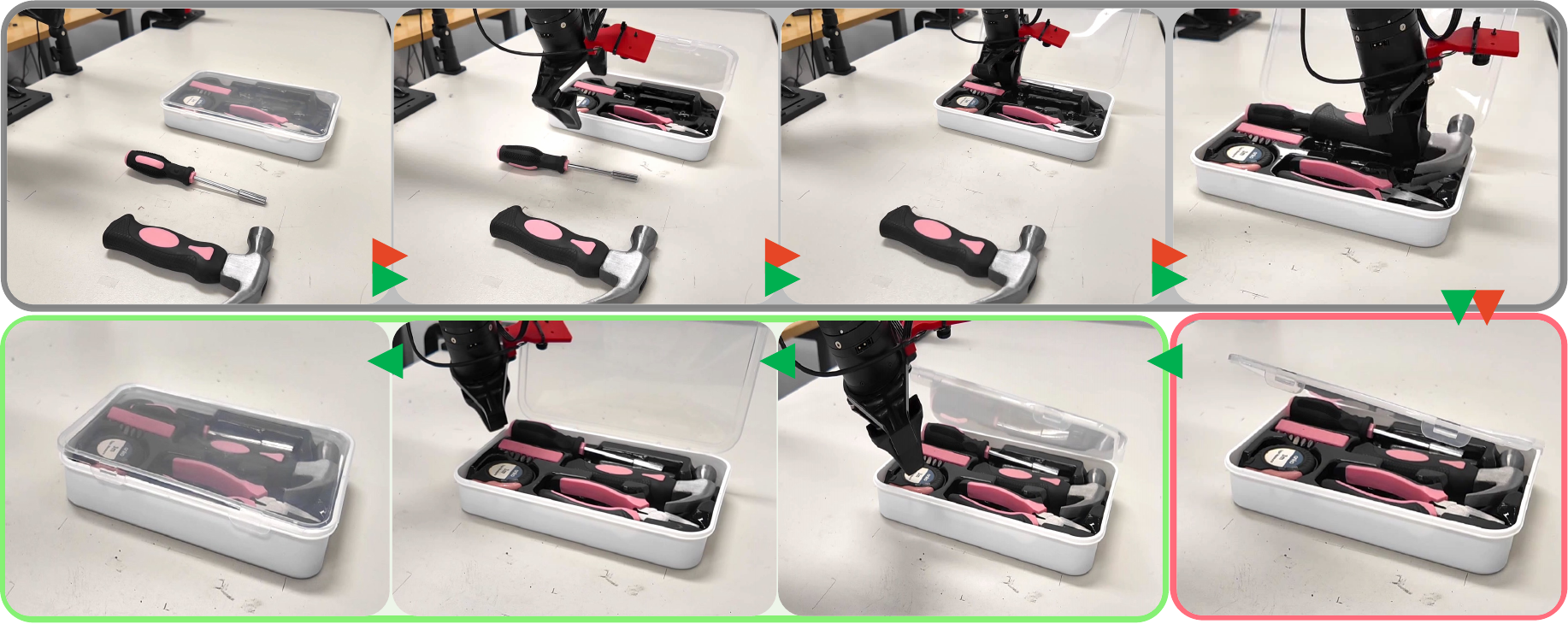}
    \caption{\textbf{Illustrative closed-loop recovery.}
    Red/green arrows show the baseline/\method{} rollout progression. The red box marks an intermediate placement failure; green boxes highlight \method{}’s subsequent corrective actions.}
    \label{fig:recovery}
\end{figure}

\vspace{0.2em}\noindent\textbf{Results:} \Cref{tab:real} shows higher \method{} success on all six task–condition pairs. The gap widens under held-out positions for Soap Pick-up and Plate Stacking, while Toolbox remains the most challenging task. \Cref{sec:shortcut_analysis} complements these behavioral results with controlled representation- and action-level input-sensitivity probes. \Cref{fig:qualitative} shows matched rollouts under nominal and shifted conditions, illustrating spatial adaptation in Soap Pick-up, fine alignment in Plate Stacking, and more precise placement in Toolbox. \Cref{fig:recovery} illustrates closed-loop recovery from an intermediate Toolbox placement error. After failed lid closure, the dense baseline remains stuck, whereas \method{} reopens the lid, corrects the placement, and completes the task.

\subsection{State-to-Image and Action Sensitivity}
\label{sec:shortcut_analysis}
We next examine whether the topological difference manifests in visual representations and action predictions by probing the compact-to-visual route through proprioception. The policy latent, at inference, is fixed to its prior mean \(z=0\)~\cite{zhao2023act}, leaving proprioception as the varying compact input; we therefore intervene on state while holding the latent fixed. Using frozen policies trained on Square, we construct state and image interventions by pairing each reference observation with a sample from another episode at the same normalized trajectory phase. State interventions replace only proprioception with the matched state, while image interventions jointly replace
the agent and wrist views with the matched images. All other inputs, including the policy latent, remain fixed. Metrics are first averaged over sampled phases within each episode, then reported as mean $\pm$ sample SD across episode means.

\vspace{0.2em}\noindent\textbf{Representation-level state imprinting:} For view $v$, let $\bb V_v^\ell\in\mathbb{R}^{N_v\times d}$ denote the nominal visual tokens at encoder depth $\ell$, and $\widetilde{\bb V}_{x,v}^\ell$ the corresponding tokens under intervention $x\in\{s,I\}$. We first measure the scale-normalized representation change: \begin{align} \delta_{x,v}^{\ell}
=
\frac{
\left\Vert
\bb V_v^\ell-\widetilde{\bb V}_{x,v}^\ell
\right\Vert_F/\sqrt{N_vd}
}{
\sigma_v^\ell+\epsilon
}, \label{eq:intervention_drift} \end{align} where $\sigma_v^\ell=\|\bb V_v^\ell\|_F/\sqrt{N_vd}$ is the activation scale of the nominal visual tokens. The normalization allows comparison across layers and views. We report $\mathbb{E}[\delta{s,v}^{\ell}]$ as the \emph{state-induced representation change}; larger values indicate stronger state influence on the spatial visual representation.

A small state-induced change could also arise from a representation that is generally insensitive to its inputs. We therefore normalize it by the corresponding image-induced change:
\begin{align}
\Gamma_v^\ell
=
\mathbb{E}\!\left[
\frac{
\delta_{s,v}^\ell
}{
\delta_{I,v}^\ell+\epsilon
}
\right].
\label{eq:relative_sensitivity}
\end{align}
The ratio is computed for each phase-matched pair before averaging. $\Gamma_v^\ell>1$ indicates greater relative sensitivity to state than image changes, whereas $\Gamma_v^\ell<1$ indicates greater relative sensitivity to the image. Agent and wrist views are reported separately here.

\begin{figure}[t]
\centering
\includegraphics[width=\linewidth]{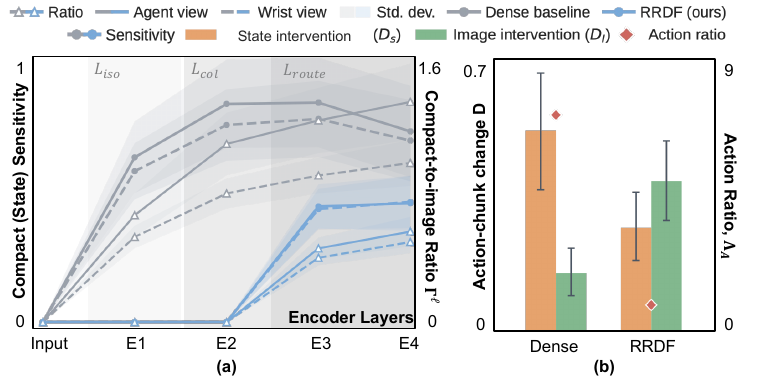}
\vspace{-2em}
\caption{\textbf{Phase-matched state-to-image and action
sensitivity analysis.}
\textbf{(a)}  State-induced visual change (circles) and
state-to-image ratio $\Gamma^\ell$ (triangles); background
bands mark RRDF's stages.
\textbf{(b)} Action changes $D_s$ and $D_I$
(bars) and ratio $\Lambda_A$ (diamonds). Shading and error
bars denote SD across episode means.}
\label{fig:shortcut}
\end{figure}

\vspace{0.2em}\noindent\textbf{Action-level intervention sensitivity:}
We additionally test whether the same imbalance appears in the policy output. Let $\widehat{\bb A}_x$ denote the predicted normalized action chunk under intervention $x\in\{s,I\}$, and define its change from the nominal prediction $\widehat{\bb A}$ as $D_x=\|\widehat{\bb A}-\widehat{\bb A}_x\|_F/\sqrt{Hd_a}$, where $H$ is the chunk horizon and $d_a$ the action dimension. We summarize the relative influence of state and image interventions as: \begin{align}
\Lambda_A
=
\mathbb{E}\!\left[
\frac{D_s}{D_I+\epsilon}
\right].
\label{eq:action_sensitivity}
\end{align}
Thus, $\Lambda_A>1$ indicates greater action-output sensitivity to state than image changes, whereas $\Lambda_A<1$ indicates greater relative sensitivity to the image.


\Cref{fig:shortcut} summarizes the intervention sensitivity analysis.  At the representation level, \method{} exhibits zero state-induced change in spatial visual tokens throughout the protected prefix, as implied by \Cref{eq:prefix_independence}. After routing begins, the two representation-level measures (the state-induced change \(\mathbb{E}[\delta_{s,v}^{\ell}]\) and state-to-image ratio \(\Gamma^\ell\)) remain below those of dense fusion at matched depths, showing that lower measured state-induced visual sensitivity persists after register-mediated readback begins. At the action level, state interventions induce substantially larger action-chunk changes than matched image interventions under dense fusion, resulting in a high \(\Lambda_A\). \method{} substantially reduces this relative state-to-image action sensitivity. Importantly, reduced state-to-image sensitivity does not remove proprioceptive conditioning: state tokens remain directly available to the action generator. Together with the task-level gains, the lower representation- and action-level state-to-image sensitivity is consistent with the motivation for controlling shortcut-prone early fusion while retaining compact action conditioning.

\subsection{Architectural Ablations}
\label{sec:ablations}
We train one policy per ablation variant using the same training seed and matched rollout initializations. \Cref{tab:ablations} groups controls into global topology, prefix composition, and register-workspace routing; per-checkpoint evaluation follows \Cref{sec:simulation}. Delayed variants use the same protected-prefix depth at 2, and ``Sep.'' denotes register-free stream separation. 



\begin{table}[t]
\centering
\caption{\textbf{Fusion-topology ablations on Square.} Nominal success rate (Nom.): the best-of-final-ten success, mean $\pm$ SD. Sensitivity statistics for $\Gamma^L$ follow \Cref{sec:shortcut_analysis}.}
\label{tab:ablations}
\footnotesize
\renewcommand{\arraystretch}{1.1}
\resizebox{\columnwidth}{!}{%
\begin{tabular}{@{}lcclcc@{}}
\toprule
Variant
& Reg.
& Prefix
& Topology
& Nom. $\uparrow$
& $\Gamma^L$ \\
\midrule

\multicolumn{6}{@{}l}{\textbf{(a) Global topology controls}} \\
\midrule
Dense
&
&
& Dense
& $75.3{\pm}3.1$
& $1.39{\pm}0.20$ \\

Dense + R
& $\checkmark$
&
& Dense
& $77.3{\pm}3.1$
& $0.96{\pm}0.29$ \\

Delayed Dense
&
& Sep.
& Dense
& $87.3{\pm}2.3$ & $0.81{\pm}0.16$ \\

R-Route
& $\checkmark$
&
& Routed
& $90.7{\pm}2.3$ & $0.39{\pm}0.01$ \\

\midrule
\multicolumn{6}{@{}l}{\textbf{(b) Pre-routing prefix decomposition}} \\
\midrule
R-Iso-Route
& $\checkmark$
& Iso.
& Routed
& $90.7{\pm}1.2$ & $0.44{\pm}0.05$ \\

R-Col-Route
& $\checkmark$
& Col.
& Routed
& $91.3{\pm}5.0$ & $\mathbf{0.21{\pm}0.02}$ \\

\midrule
\multicolumn{6}{@{}l}{\textbf{(c) Register-workspace routing}} \\
\midrule
R-Shared
& $\checkmark$
& Shared
& Routed
& $88.7{\pm}6.4$
& $0.58{\pm}0.14$ \\

R-Coupled
& $\checkmark$
& Coupled$\rightarrow$Col.
& Routed
& $91.3{\pm}5.8$
& $0.42{\pm}0.07$ \\

R-RestrCol
& $\checkmark$
& Iso.$\rightarrow$Restr. Col.
& Routed
& $92.7{\pm}4.2$
& $0.47{\pm}0.10$ \\

\midrule
\textbf{\method{}}
& $\checkmark$
& Iso.$\rightarrow$Col.
& Routed
& $\mathbf{93.3{\pm}3.1}$
& $0.45{\pm}0.11$ \\

\bottomrule
\end{tabular}%
}
\end{table}

\vspace{0.2em}\noindent\textbf{Fusion-topology decomposition:}
Adding registers to dense fusion yields only a small performance change, indicating that additional token capacity alone does not explain the gain.
Delaying direct compact--visual fusion without registers produces a much larger improvement, showing that fusion onset is itself an important architectural factor. The register-mediated variants consistently achieve lower sensitivity, $\Gamma^L$, while improving performance.

Both isolation-only and collection-only prefixes improve over dense fusion. The full schedule achieves the highest mean success, although simplified schedules perform similarly. The collection-only prefix produces the lowest state sensitivity without the highest success, indicating that minimum sensitivity is not a sufficient model-selection criterion.

\vspace{0.2em}\noindent\textbf{Register-workspace routing:} Compared with R-Shared, which lets all registers read both primary streams during Isolation, source-specific access improves task performance and reduces compact-to-image sensitivity. R-Coupled introduces cross-register interaction during Isolation, whereas R-RestrCol maintains source-specific access through Collection; both retain most of the gain. The full \method{} schedule achieves the highest mean success by combining source-specific accumulation with joint collection.

\begin{figure}
    \centering
    \includegraphics[width=\linewidth]{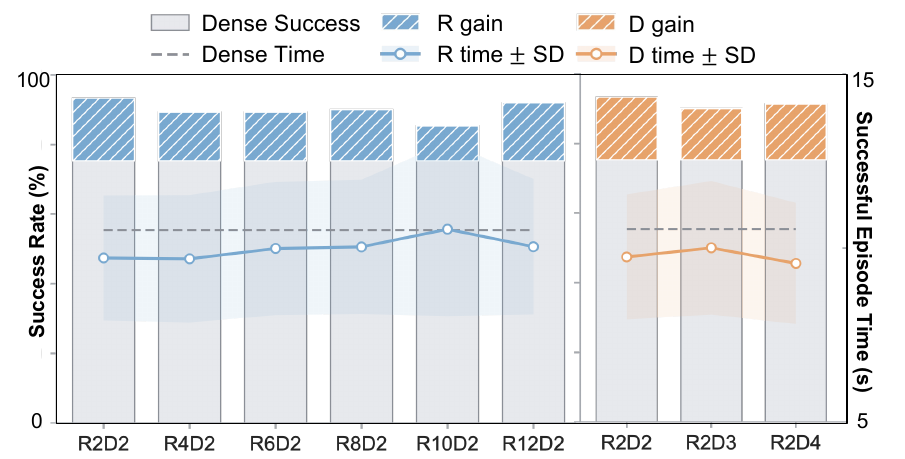}
\caption{\textbf{Sensitivity to workspace width and routing onset on Square.} R$i$D$j$ denotes $R=i$ registers and $L_{\rm delay}=j$, with $L=4$; D2/D3/D4 use
$(L_{\rm iso},L_{\rm col},L_{\rm route})=(1,1,2)/(1,2,1)/(1,3,0)$. Left: register-count sweep with $L_{\rm delay}=2$; right: routing-delay sweep with $R=2$.
D4 is the no-encoder-readback boundary case. Bars show success and gains over dense fusion; lines and shading show mean completion time $\pm$ SD.} \vspace{0.5em}
\label{fig:ablation_sensitivity}
\end{figure}

\vspace{0.2em}\noindent\textbf{Register count and routing delay:}
\Cref{fig:ablation_sensitivity} examines the two structural hyperparameters of \method{}, reporting the best-of-final-ten results. Varying the register count $R$ reveals a broad high-performing region around $R=2$--$6$, with no consistent benefit from further increasing register workspace width, indicating weak sensitivity to register count within this range. The default R2D2 schedule achieves the highest observed success, while gains over dense fusion persist across the tested readback settings. This includes the no-encoder-readback endpoint R2D4. We retain R2D2 as the default because it combines an early protected prefix with later register-mediated feedback.

\begin{table}[t]
\centering
\caption{\textbf{Visual-backbone ablation on Square.} Best-of-final-ten results, mean $\pm$ SD.}
\label{tab:visual_backbone_ablation}
\footnotesize
\renewcommand{\arraystretch}{1.1}
\setlength{\tabcolsep}{3pt}
\begin{tabular}{@{}lcccc@{}}
\toprule
& \multicolumn{2}{c}{Dense ACT}
& \multicolumn{2}{c}{\method} \\
\cmidrule(lr){2-3}
\cmidrule(l){4-5}
Visual backbone
& SR (\%) $\uparrow$
& Time (s) $\downarrow$
& SR (\%) $\uparrow$
& Time (s) $\downarrow$ \\
\midrule
ResNet-18
& $75.3 \pm 3.1$
& $10.56 \pm 0.68 $
& $\mathbf{93.3 \pm 3.1}$
& $9.74 \pm 0.25$ \\

DINOv2-Base
& $76.7 \pm 9.0$
& $9.89 \pm 2.25$
& $\mathbf{88.0 \pm 4.0}$
& $10.04 \pm 2.25$ \\
\bottomrule
\end{tabular}
\end{table}

\begin{table}[h]
\centering
\caption{\textbf{Diffusion Policy transfer on Square.} Success and successful-rollout time: mean ± sample SD across the final five checkpoint means.}
\label{tab:dp_result}
\footnotesize
\renewcommand{\arraystretch}{1.08}
\setlength{\tabcolsep}{6pt}
\begin{tabular}{@{}lcc@{}}
\toprule
Fusion & SR (\%) $\uparrow$ & Time (s) $\downarrow$ \\
\midrule
Feature concat & $81.1 \pm 3.8$ & $7.50 \pm 0.27$ \\
Dense fusion & $80.8 \pm 3.9$ & $7.19 \pm 0.16$ \\
\method{} & $\mathbf{84.3 \pm 3.3}$ & $7.26 \pm 0.08$ \\
\bottomrule
\end{tabular}
\end{table}

\vspace{0.2em}\noindent\textbf{Visual backbone and action-generator transfer:}
On Square, we replace ResNet-18 with DINOv2-Base~\cite{oquab2023dinov2} and compare transformer-based Diffusion Policy’s~\cite{chi2023diffusionpolicy} native feature concatenation with dense spatial-token fusion and RRDF. Both comparisons favor \method{} (\Cref{tab:visual_backbone_ablation},\Cref{tab:dp_result}), extending the evaluation beyond default ACT.

\section{Conclusion}
\label{sec:conclusion}
We introduced Register-Routed Delayed Fusion (\method{}), a source-partitioned attention schedule that masks direct compact-visual exchange and delays register-mediated feedback while preserving compact action conditioning. Across the evaluated simulation and real-robot tasks, \method{} matches or improves nominal performance and achieves higher success under the tested appearance and position shifts. Ablations favor structured fusion over registers alone, while phase-matched probes reveal lower measured state-to-image sensitivity. Results with backbone and action-generation transfer support compatibility beyond the default ACT configuration, highlighting \method{}'s flexibility as a modular observation-fusion component for visuomotor policies.

\bibliographystyle{IEEEtran}
\bibliography{ref}

\end{document}